# Efficient Parallel Estimation for Markov Random Fields

Michael J. Swain, Lambert E. Wixson and Paul B. Chou *

Computer Science Department
University of Rochester
Rochester, NY 14627

## Abstract

We present a new, deterministic, distributed MAP estimation algorithm for Markov Random Fields called Local Highest Confidence First (Local HCF). The algorithm has been applied to segmentation problems in computer vision and its performance compared with stochastic algorithms. The experiments show that Local HCF finds better estimates than stochastic algorithms with much less computation.

## 1 Introduction

The problem of assigning labels from a fixed set to each member of a set of sites appears at all levels of computer vision. Recently, an optimization algorithm known as Highest Confidence First (HCF) [Chou, 1988] has been applied to labeling tasks in low-level vision. Examples of such tasks include edge detection, in which each inter-pixel site must be labeled as either edge or non-edge, and the integration of intensity and sparse depth data for the labeling of depth discontinuities and the generation of dense depth estimates. In these tasks, it often outperforms conventional optimization techniques such as simulated annealing[Geman and Geman, 1984], Monte Carlo sampling[Marroquin et al., 1985], and Iterative Conditional Modes (ICM) estimation[Besag, 1986].

The HCF algorithm is serial, deterministic, and guaranteed to terminate. We have developed a parallel version of HCF, called Local HCF, suitable for a SIMD architecture in which each processor must only communicate with a small number of its neighbors. Such an architecture would be capable of labeling an image in real-time. Experiments have shown that Local HCF almost always performs better than HCF and much better than the techniques just mentioned.

In the next section, the labeling problem is discussed. Sections 3 and 4 review Markov Random Fields and Chou's HCF algorithm. Section 5 describes the Local HCF algorithm, and test results are presented in Section 6. Finally, we discuss future plans for this research.

## 2 Generating Most Probable Labelings

In probabilistic labeling, *a priori* knowledge of the frequency of various labelings and combinations of labelings can be combined with observations to find the *a posteriori* probabilities that each site should have a certain label. For complexity reasons, the interactions among the variables are usually modeled as Markov Random Fields, in which a variable interacts with a restricted number of other variables called neighbors. If a link is drawn between all neighboring variables the resulting graph is called the neighborhood graph.

The problem is to find the labeling which has the highest probability given the input data. This is called the maximum a posteriori (MAP) labeling.[1] For a Markov Random Field, the MAP labeling can be found by locating the minimum of the Gibbs energy, which is a function of both *a priori* knowledge (expressed as energies associated with cliques in the neighborhood graph) and the input data.

A major problem with the probabilistic labeling approach is the exponential complexity of finding the exact MAP estimate. The methods mentioned in Section 1 have traditionally been used to find labelings whose energies are close to the global min-imum of the energy function. Simulated annealing has been widely used for this purpose because of the elegant convergence proofs associated with the algorithm and its massively parallel nature [Geman and Geman, 1984]. But in practice simulated annealing is slow and highly dependent on the cooling schedule and initial configuration. Marroquin's Monte Carlo MPM estimator has much better performance in practice, although its performance is also dependent on the initial configuration. A bad initial configuration slows or prevents the algorithm from reaching a good configuration (see Figure 7). Continuation methods have also been used for specific classes of computer vision reconstruction problems formulated as Markov Random Fields [Koch et al., 1986; Blake and Zisserman, 1987], but these cannot be applied to arbitrary MRF estimation problems. Although Koch's approach is efficient in analogue VLSI, it is slow to simulate on a grid of standard processors. Hummel and Zucker's [1983] relaxation labeling technique could also be applied to the MRF estimation problem, but it does not guarantee generating a *feasible* solution, that is, one in the space of possible solutions.

---

*Current address: IBM T.J. Watson Research Center, P. O. Box 704, Yorktown Heights, NY 10598

[1][Marroquin et al., 1985] points out that the Maximizer of Posterior Marginals is more useful when the data is very noisy. This labeling minimizes the expected number of mislabeled sites. Marroquin uses a Monte Carlo procedure to compute this MPM labeling. Unless the Monte Carlo procedure is given a very good initial estimate, HCF produces better segmentations.

Faced with the problems posed by traditional optimization methods, Chou developed an algorithm called Highest Confidence First (HCF). Unlike the stochastic energy minimization procedures, the HCF algorithm is *deterministic* and *guaranteed to terminate* at a local minimum of the energy function. One drawback of HCF is that it is a serial algorithm. This discourages its real-time application to problems with large numbers of sites, and would also cast doubt on any hypothesized connection between HCF and biological plausibility. This paper presents a parallel adaptation of HCF, called Local HCF.

## 3 Markov Random Fields

A Markov Random Field is a collection of random variables $S$ which has the following locality property:

$$P(X_s = \omega_s | X_r = \omega_r, r \in S, r \neq s) = $$
$$P(X_s = \omega_s | X_r = \omega_r, r \in N_s, r \neq s)$$

where $N_s$ is known as the neighborhood of the random variable $X_s$. The MRF is associated with an undirected graph called the neighborhood graph in which the vertices represent the random variables. Vertices are adjacent in the neighborhood graph if the variables are neighbors in the MRF.

Denote an assignment of labels to the random variables by $\omega$. The Hammersley-Clifford theorem [Besag, 1974] shows that the joint distribution $P(\omega)$ can be expressed as a normalized product of positive values associated with the cliques of the neighborhood graph. This can be written:

$$P(\omega) = \frac{e^{-U}}{Z}$$

where

$$U = \sum_{c \in C} V_c(\omega)$$

and Z is a normalizing constant. The value $U$ is referred to as the energy of the field; minimizing $U$ is equivalent to maximizing $P(\omega)$. In this notation, the positivity of the clique values is enforced by the exponential term and the clique parameters $V_c$ may take on either positive or negative values.

Normally, the unary clique values are broken into separate components representing prior expectations $P(\omega_s)$ and liklihood values obtained from the observations $P(O_s|\omega_s)$. This is done using Bayes rule, which states

$$P(\omega_s|O_s) = \frac{P(\omega_s)P(O_s|\omega_s)}{P(O_s)}. \qquad (1)$$

If the $V_c$'s are used to signify the prior expectations, then $U$ is revised to read:

$$U = \sum_{c \in C} V_c(\omega) - \sum_{s \in S} \log P(O_s|\omega_s)$$

The denominator in Equation 1 is absorbed into the normalizing constant $Z$.

## 4 HCF

In the HCF algorithm all sites initially are specially labeled as "uncommitted", instead of starting with some specific labeling as with previous optimization methods. Cliques for which any member is uncommitted do not participate in the computation of the energy of the field. For each site, a stability measure is computed. The more negative the stability, the more confidence we have in changing its labeling. On each iteration, the site with minimum stability is selected and its label is changed to the one which creates the lowest energy. This in turn causes the stabilities of the site's neighbors to change. The process is repeated until all changes in the labeling would result in an increase in the energy, at which point the energy is at local minimum in the energy function and the algorithm terminates. The algorithm is given in Figure 1.

The stability of a site is defined in terms of a quantity known as the *augmented a posteriori local energy*

```
begin
    ω = (l₀, ..., l₀);
    top = Create_Heap(ω);
    while stability_top < 0 do
        s = top;
        Change_State(ω_s);
        Update_Stability(stability_s);
        Adjust_Heap(s);
        for r ∈ N_s do
            Update_Stability(stability_r);
            Adjust_Heap(r)
        end
    end
end
```

Figure 1: The algorithm HCF

$E$, which is:

$$E_s(l) = \sum_{c:s \in c} V'_c(\omega') - \sum_{s \in S} \log P(O_s|\omega_s)$$

where $\omega'$ is the configuration that agrees with $\omega$ everywhere except that $\omega'_s = l$. Also, $V'_c$ is 0 if $\omega_r = l_0$, the uncommitted state, for any $r$ in $c$, otherwise it is equal to $V_c$.

The stability $G$ of an uncommitted site $s$ is the negative difference between the two lowest energy states that can be reached by changing its label:

$$G_s(\omega) = -min_{k \in L, k \neq \omega_{min}}(E_s(k) - E_s(\omega_{min}))$$

In this expression $\omega_{min} = \{k|E_k \text{ is a minimum}\}$. The stability of a committed site is the difference between it and the lowest energy state different from the current state $\omega_s$:

$$G_s(\omega) = min_{k \in L, k \neq \omega_s}(E_s(k) - E_s(\omega_s)).$$

## 5 Local HCF

The Local HCF algorithm is a simple extension of HCF: *On each iteration, change the state of each site whose stability is negative and less than the stabilities of its neighbors.* In a preprocessing phase, the sites are each given a distinct rank, and, if two stabilities are equal in value, the site with lower rank is considered to have lower stability. These state changes are done in parallel, as is the recalculation of the stabilities for each site. The algorithm terminates when no states are changed. Pseudocode for Local HCF is given in Figure 2, for which you should assume a processor is assigned to every element of the *site* data

```
site: parallel array[1..N_SITES] of record
    stability;
    i; /* rank */
    change;
end

begin
    with site do in parallel
        do
            begin
                change := false;
                Update_Stability(stability);
                (nbhd_stability,k) := min_{n ∈ N[i]} (site[n].stability,site[n
                if stability < 0 and (stability,i) < (nbhd_stability,k)
                    begin
                        Change_State(state);
                        change := true;
                    end
                any_change := (&all change);
            until any_change = false;
        end
end
```

Figure 2: The algorithm Local HCF

structure. The algorithm is written in a notation similar to C* [Rose and Steele, 1987], a programming language developed for the Connection Machine. In the algorithm, the operator &all returns the result of a global *and* operation.

For the low-level vision tasks which we have studied, the MRFs have uniform spatial connectivity and uniform clique potential functions. Thus, Local HCF applied to these tasks is well suited for a massively parallel SIMD approach which assigns a simple processor to each site. Each processor need only be able to examine the states and stabilities of its neighbors. The testing and updating of the labels of each site can then be executed in parallel. Such a neighborhood interconnection scheme is simple, cheap, and efficient.

Like HCF, Local HCF is deterministic and guaranteed to terminate. It will terminate because the energy of the system decreases on each iteration. We know this because (roughly) (a) at least one site changes state per iteration — there is always a site whose stability is a minimum — and (b) the energy change per iteration is equal to the sum of the stabilities of the sites which are changed. These stabilities are negative and the state changes will not interact with each other because none of the changed sites are neighbors. Therefore, the energy of the system always decreases. A rigorous proof of convergence is given in Appendix A.

Determinism and guaranteed termination are valuable features. Analysis of results is much easier; for each set of parameters, only one run is needed to evaluate the performance, as opposed to a sampling of runs, as with simulated annealing.

## 6 Test Results

We have chosen to use edge detection as our test domain. In this task, each site is either vertical or horizontal and appears between two pixels. The problem is to label each site as either an *edge* or a *non-edge*, based on the intensities of the pixels.

We have added an implementation of Local HCF to the simulator originally constructed by Chou, which allows us to compare the final labelings produced by Local HCF, HCF, and a variety of standard labeling techniques. The simulator runs on a Sun workstation.

The input is produced by Sher's probabilistic edge detector [Sher, 1987] and consists of the log likelihood ratio for an edge at each site. The algorithms were tested on likelihood ratios from the checkerboard image, the "P" block image, and the "URCS" block image which appear in Figure 5. As a much harder test, the algorithms were also presented with noisy (corrupted) likelihood ratios obtained by using an incomplete edge model to find edges in the "URCS" image.

The clique energies were chosen in an *ad hoc* manner. They were chosen to encourage the growth of continuous line segments and to discourage abrupt breaks in line segments, close parallel lines (competitions) and sharp turns in line segments. "Encouragement" or "discouragement" is associated with a clique by assigning it a negative or positive energy, respectively. To encode these relationships, a second-order neighborhood, in which each site is adjacent to eight others, is used. This neighborhood system is shown in Figure 3 and the clique values used are shown in Figure 4.

The goodness of the result of applying one of the labeling algorithms can be determined qualitatively by simply looking at a picture of the segmentation, and quantitatively by examining the energy of the final configuration. Figure 6 shows the labelings produced by Local HCF on the four test cases, and Figures 8–10 compare the algorithms over time.

Figure 7 shows energies of the final configurations yielded by thresholding the likelihood ratio of edge to non-edge (TLR), simulated annealing MAP estimation, Monte Carlo MPM estimation, ICM estimation (scan-line order), ICM (random order), HCF, and our results from Local HCF. The values of MAP, MPM, and the ICM's are the averages of the results from several runs. In almost every case, Local HCF found the labeling with the least energy. Each Local HCF run took 20-30 iterations (parallel state changes); we expect that the Connection Machine will carry out these labelings almost instantaneously.

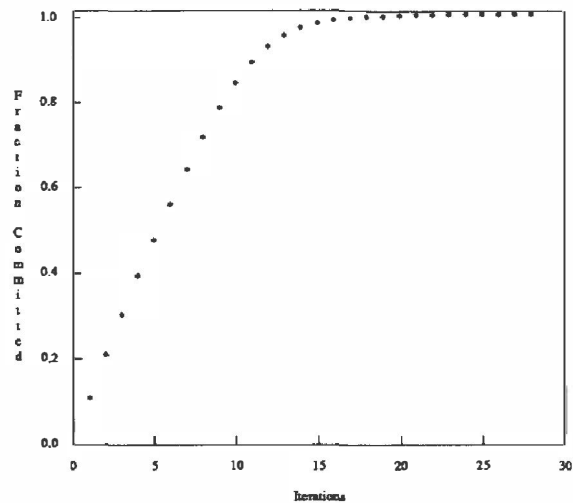

Figure 8: Fraction of sites committed versus iterations of Local HCF. Most sites commit early in the computation.

We believe that Local HCF performs better than HCF because it is much less likely to propagate the results of local labelings globally across the image. The execution of HCF is often marked by one site $s$ committing to a certain label, immediately followed by one of its neighbors $s + 1$ committing to a label which is compatible with the new label of $s$. This process is then repeated for a neighbor of $s + 1$, and its neighbor, and so on. In this manner, the effects of locally high confidence can get propagated too far. Local HCF does not tend to propagate information as far. Appendix B develops this argument in more detail.

## 7 Conclusions and Future Work

We have introduced a parallel labeling algorithm for Markov Random Fields which produces better labelings than traditional techniques at a much lower computational cost. Empirically, ten iterations on a locally connected parallel computer is sufficient to almost completely label an entire image; forty itera-

364

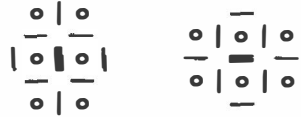

Figure 3: Neighborhoods for vertical and horizontal edge sites. Circles represent pixels, the thick line represents the site, and thin lines represent the neighbors of the site.

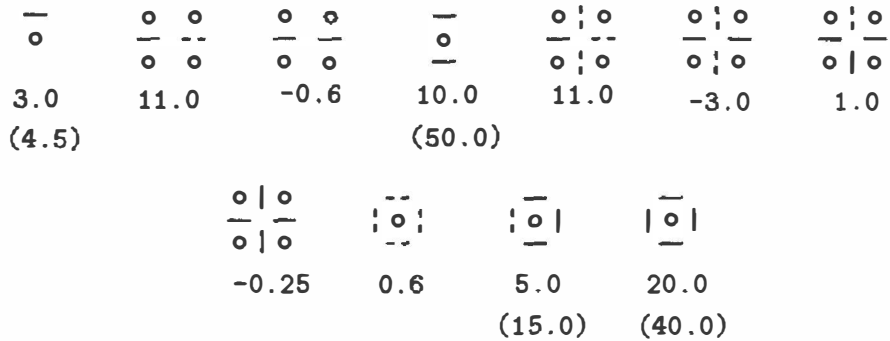

Figure 4: Clique energies. The parenthesized values were used for the corrupted edge data (see Figure 6).

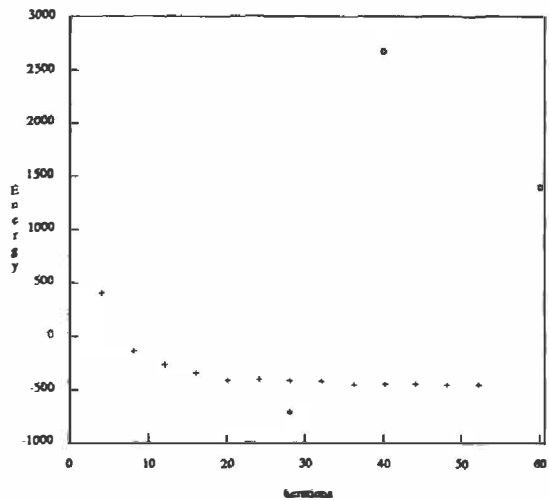
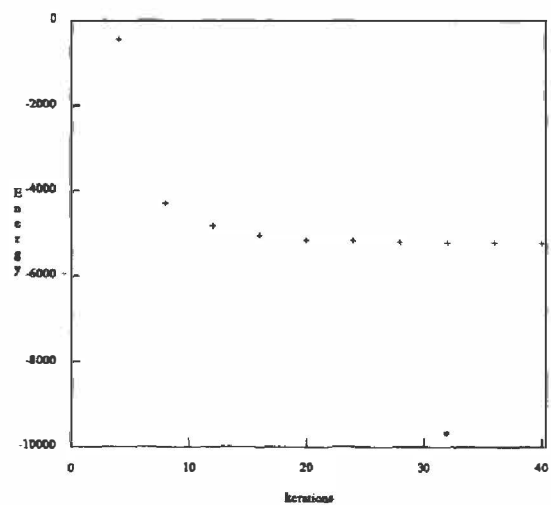

Figure 9: Timecourse of parallel algorithms on the URCS image. * = Local HCF, + = Monte Carlo MPM, o = Simulated Annealing.

Figure 10: Timecourse of parallel algorithms on the corrupted URCS image. * = Local HCF, + = Monte Carlo MPM.

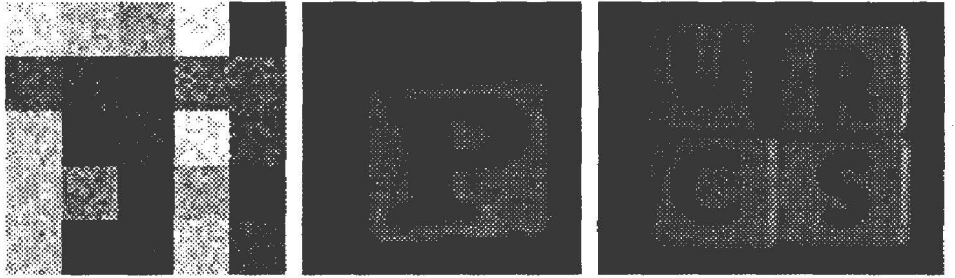

Figure 5: Test images (8 bits/pixel). Checkerboard and P images are 50 × 50. URCS image is 100 × 124.

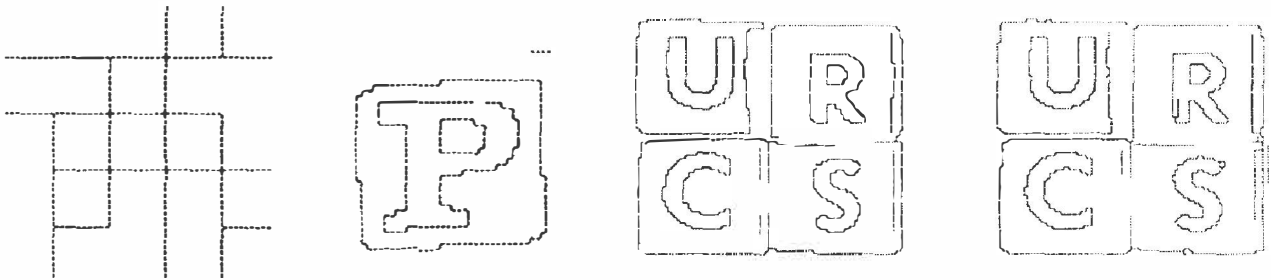

Figure 6: Edge labelings produced by Local HCF for above images. Rightmost labeling demonstrates Local HCF on noisy edge data from the URCS image.

| Method | Checkerboard image | P image | URCS image | Corrupted URCS image |
|---|---|---|---|---|
| TLR | -3952 | -572 | 4785 | 59719 |
| Annealing | -4282 | -680 | -349 | -5303 |
| MPM | -4392 | -723 | -503 | -5296 |
| ICM(s) | -4364 | -693 | -503 | -4954 |
| ICM(r) | -4334 | -715 | -513 | -3728 |
| HCF | -4392 | -750 | -380 | -9635 |
| Local HCF | -4392 | -720 | -625 | -9648 |

Figure 7: Energy Values. (The smaller the energy the closer the labeling is to the MAP estimate.)

tions finishes the task. In the future we intend to study its applicability in situations in which there are much larger number of labels, such as occurs in recognition problems [Cooper and Swain, 1988]. We are also studying an extension of Pearl's method for determining clique parameters on chordal graphs [Pearl, 1988] to more general graphs.

## A Proof of Convergence for Local HCF

We prove that the algorithm terminates, and returns a feasible solution which is at a local minimum of the energy function.

Define the ordered stability of a site to be a pair $(a, b)$ where $a$ is the stability and $b$ is the rank of the site. Then $(a, b) < (c, d)$ iff

1. $a < c$ or

2. $a = c$ and $b < d$.

**Lemma 1** *For at least one site $k$, the ordered stability, denoted $s^*$, is a minimum in its neighborhood. That is:*

$$s_k^* < \min_{n \in N[i]} s_n^*.$$

**Proof:** Since the ordered stability value imposes a strict ordering on the sites in the Markov Random Field, there is one site with the minimum ordered stability in the field. That site will also have the minimum ordered stability in its neighborhood. □

It is straightforward to design cases in which the stabilities are ordered so that only one site is a minimum in its neighborhood. This is done by creating a field in which the stability has only one local minimum, which is also the global minimum. Therefore, parallelism is not guaranteed, but empirically the running time is nearly independent of the number of variables in the random field.

**Lemma 2** *At each iteration of Local HCF either*

*1. The number of committed sites increases, or*

*2. The energy function decreases.*

**Proof:** Sites may only commit once, so the number of committed sites may not decrease. We consider the case when the number of committed sites stays constant. If two sites are members of the same clique then each is a neighbor of the other, so no two sites in the same clique change state in one iteration of Local HCF. Therefore, the change in energy is the sum of the stabilities of the sites that changed state. Each of the stabilities is negative, and so their sum is also negative. □

**Theorem 1** *Local HCF terminates at a local minimum of the energy function.*

**Proof.** By Lemma 2 the set of sites cannot return to the same state. Therefore, since there are only a finite number of configurations, Local HCF must terminate. At termination, all of the stabilities of the sites are non-negative, and so the final configuration is in the space of possible solutions because the stability of uncommitted sites are all negative. Changing the commitment of any single site cannot decrease the energy function, and so the final configuration is a local minimum of the energy function. □

## B Comparing Local HCF and HCF

Local HCF empirically produces labelings that are as good or better than those found by HCF. When Local HCF produces better labelings, it is usually because HCF has propagated strong local information too far. HCF exhibits a horizon effect, that is, puts off making 'unpleasant' choices in a similar way to game playing programs with limited look-ahead. Because of this effect, labelings produced by HCF can to be overly influenced by sites chosen early in the computation. A simple one-dimensional example exhibits the problem with HCF and shows how Local HCF avoids it.

Consider a linear array of variables representing edge (e) or non-edge (n), each neighbors with the two adjacent variables. The neighborhood graph is then a chain containing unary and binary cliques. Assign values to the cliques as follows:

> Unary cliques edge, non-edge: 0 0 (edge, non-edge equally likely)
>
> Binary cliques (e,e) (n,n) (e,n) : -0.5 1 -0.5 (line breaks discouraged)

Suppose an edge detector reports the following log liklihood ratios, log ( P(observation | edge) / P(observation | non-edge) ):

4 -0.2 -0.4 -0.5 -0.3 0.1 -0.3 -0.4

By a dynamic programming method [Blake *et al.*, 1986] the optimal labeling can be found to be

e n n n n n n n

HCF produces the labeling

e e e e e e e e

because it is always locally favorable to extend the edge labeling that was initiated by the strong evidence at the left hand variable than to introduce a line break. HCF would propagate the line indefinitely, given continued weak evidence against an edge. Local HCF produces the optimal labeling, because sites of locally minimum stability commit to non-edge before the evidence from the left-hand variable propagates across the entire field.